\documentclass[letterpaper]{article} 
\usepackage{aaai24}  
\usepackage{times}  
\usepackage{helvet}  
\usepackage{courier}  
\usepackage[hyphens]{url}  
\usepackage{graphicx} 
\usepackage{natbib}  
\usepackage{caption} 
\usepackage{algorithm}
\usepackage{algorithmic}
\usepackage{amsmath}
\usepackage{amssymb}
\usepackage{subcaption}
\usepackage{multirow}
\usepackage{xcolor}
\usepackage{pifont}
\usepackage{newfloat}
\usepackage{listings}

\usepackage{color,soul}
\usepackage{xcolor}

\DeclareCaptionStyle{ruled}{labelfont=normalfont,labelsep=colon,strut=off} 
\floatstyle{ruled}
\newfloat{listing}{tb}{lst}{}
\floatname{listing}{Listing}
\title{Semantic Segmentation in Multiple Adverse Weather Conditions\\ with Domain Knowledge Retention}
\author{
	Xin Yang\textsuperscript{\rm 1}, 
	Wending Yan\textsuperscript{\rm 2},
	Yuan Yuan\textsuperscript{\rm 2},
	Michael Bi Mi\textsuperscript{\rm 2}, 
	Robby T. Tan\textsuperscript{\rm 1}
}
\affiliations{
	\textsuperscript{\rm 1}National University of Singapore \\
	\textsuperscript{\rm 2}Huawei International Pte Ltd\\
	
	e0674612@u.nus.edu, \{yan.wending,yuanyuan10@huawei\}.com, michaelbimi@yahoo.com, robby.tan@nus.edu.sg
}

\begin{document}
\maketitle

\begin{abstract}
Semantic segmentation's performance is often compromised when applied to unlabeled adverse weather conditions.
Unsupervised domain adaptation is a potential approach to enhancing the model's adaptability and robustness to adverse weather.
However, existing methods encounter difficulties when sequentially adapting the model to multiple unlabeled adverse weather conditions.
They struggle to acquire new knowledge while also retaining previously learned knowledge.
To address these problems, we propose a semantic segmentation method for multiple adverse weather conditions that incorporates adaptive knowledge acquisition, pseudo-label blending, and weather composition replay.
Our adaptive knowledge acquisition enables the model to avoid learning from extreme images that could potentially cause the model to forget.
In our approach of blending pseudo-labels, we not only utilize the current model but also integrate the previously learned model into the ongoing learning process. 
This collaboration between the current teacher and the previous model enhances the robustness of the pseudo-labels for the current target.
Our weather composition replay mechanism allows the model to continuously refine its previously learned weather information while simultaneously learning from the new target domain.
Our method consistently outperforms the state-of-the-art methods, and obtains the best performance with averaged mIoU (\%) of {\bf 65.7} and the lowest forgetting (\%) of {\bf 3.6} against 60.1 and 11.3 \cite{hoyer2023mic}, on the ACDC datsets for a four-target continual multi-target domain adaptation.
\end{abstract}

\section{Introduction}

\begin{figure}[ht!]
	\centering
	\centering
	\begin{subfigure}[b]{\columnwidth}
		\includegraphics[width=0.242\columnwidth,height=0.15\linewidth]{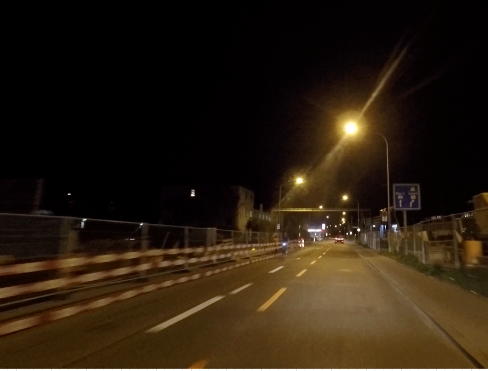}
		\includegraphics[width=0.242\columnwidth,height=0.15\linewidth]{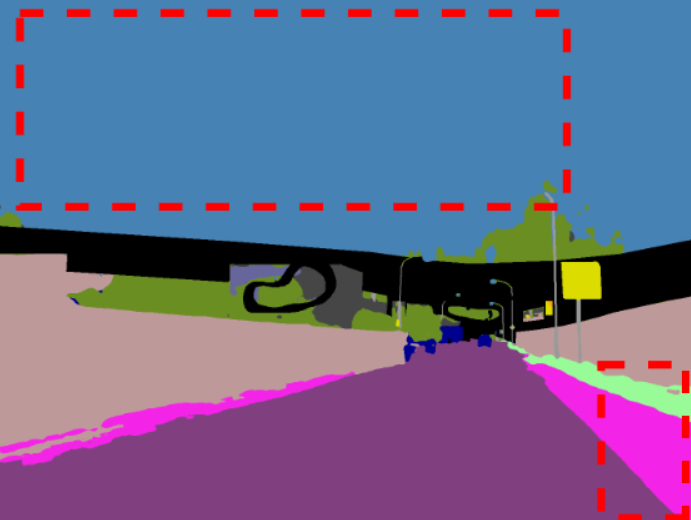}
		\includegraphics[width=0.242\columnwidth,height=0.15\linewidth]{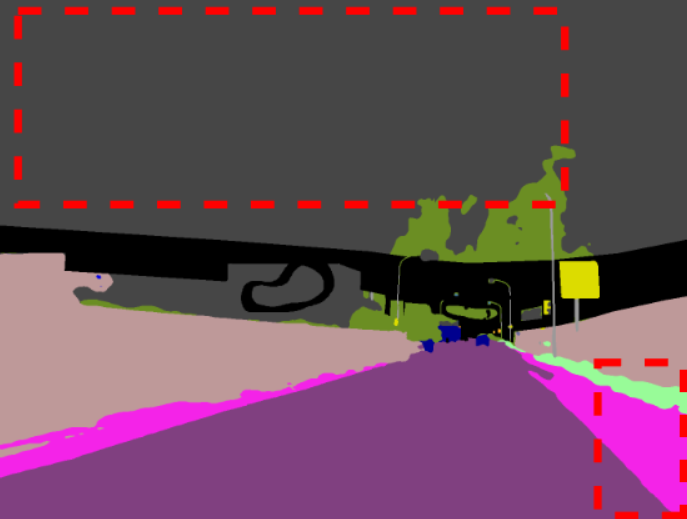}
		\includegraphics[width=0.242\columnwidth,height=0.15\linewidth]{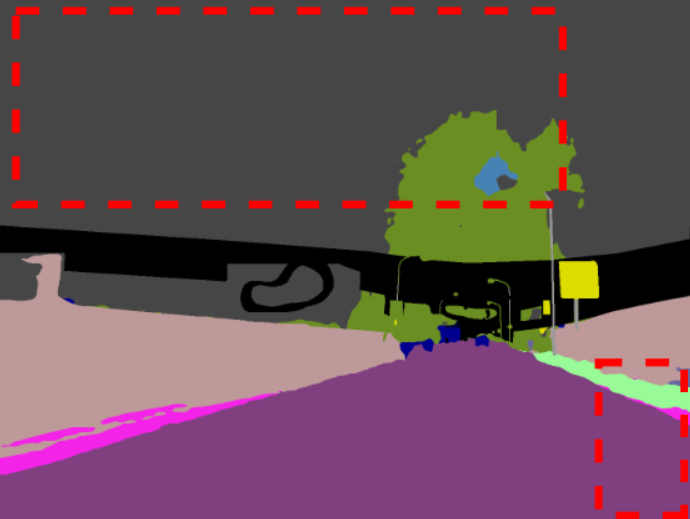} 
		\caption*{\ \  \ \  \ \ \ \    Image   \ \ \ \ \ \ \ \ \ \ \ \    \    MIC  on T1 \ \   $\rightarrow$  \ \     T1 after T2 \    $\rightarrow$   \  T1 after T3 \ \ \ \  }
	\end{subfigure}
	\\
	
	\centering
	\begin{subfigure}[b]{\columnwidth}
		\centering
		\includegraphics[width=0.242\columnwidth,height=0.15\linewidth]{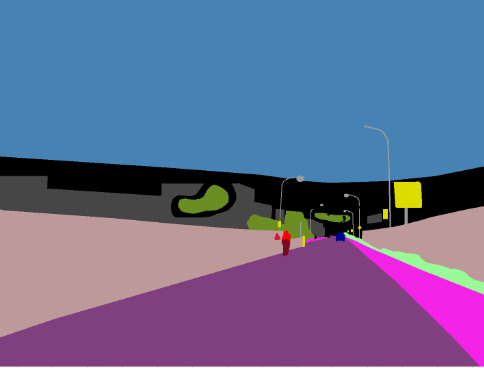}  
		\includegraphics[width=0.242\columnwidth,height=0.15\linewidth]{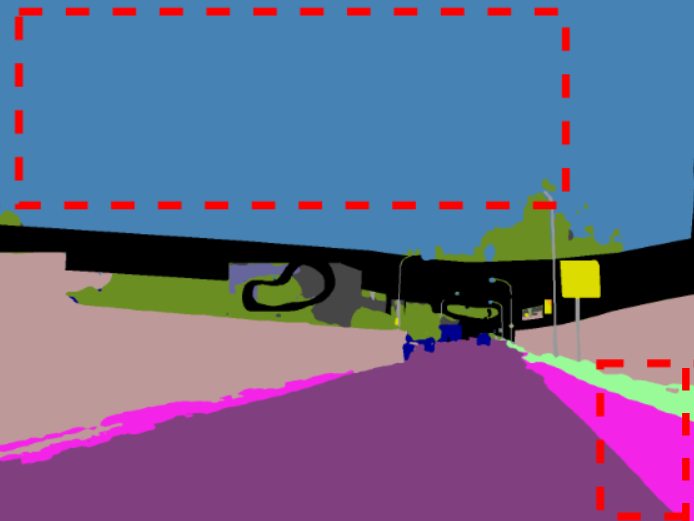}  
		\includegraphics[width=0.242\columnwidth,height=0.15\linewidth]{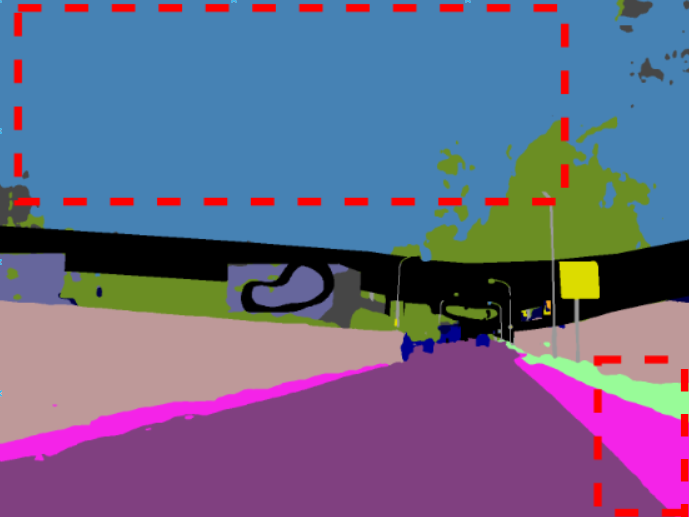} 
		\includegraphics[width=0.242\columnwidth,height=0.15\linewidth]{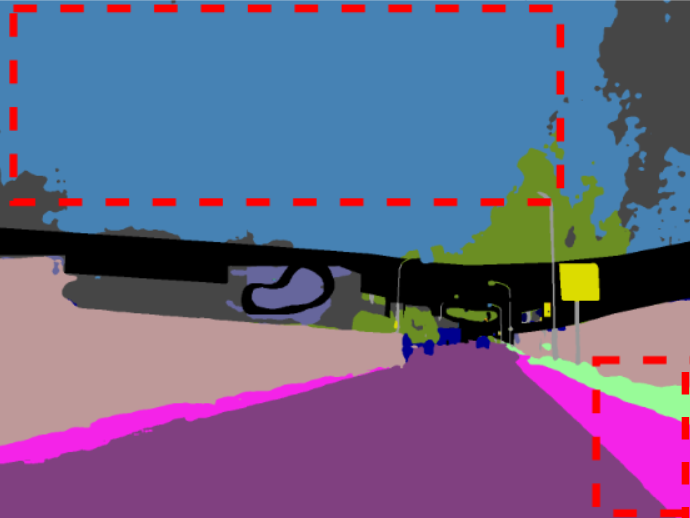} 
		\caption*{ \ \ Ground Truth \textbf{  \ \  \ \ Ours on T1 \  $\rightarrow$ \     T1 after T2\ \   $\rightarrow$ \    T1 after T3\ \  \  }}
	\end{subfigure}
	\caption{\label{fig:proc} 
	The illustration of our method, where the model adapts to each target domain sequentially. MIC \cite{hoyer2023mic} fails to retain previously learned knowledge, as its performance on the first target gradually deteriorates, e.g., the sky and the side walk in Target 1 are disappearing after the method learns Targets 2 and 3. Our method retain previously learned knowledge while adapting to new targets.
	}
\end{figure}

Semantic segmentation methods face challenges in adverse weather conditions, as these conditions significantly degrade images.
Adapting models to multiple adverse weather conditions in an unsupervised and successive manner causes additional difficulties due to substantial domain gaps among various weather conditions, and potentially leads to forgetting previously learned knowledge.

Continual unsupervised domain adaptation emerges as a potential solution to address these challenges by adapting the model from the labeled source domain to the unlabeled target domains in a sequential manner, e.g., (\cite{lin2022prototype,saporta2022multi}).
However, these methods are designed to acquire all information from the new target, without considering whether this information might lead to a forgetting of previously learned knowledge.
Moreover, as mentioned in \cite{kalb2023principles}, domain shifts across distinct adverse weather conditions are primarily induced by the deteriorated information within low-level features extracted from the early convolutional layers. Consequently, there is a necessity to develop a method that takes this factor into consideration to effectively address the challenges posed by adverse weather conditions in domain adaptation.

In this paper, we present a semantic segmentation method that sequentially adapts the model to multiple unlabeled adverse weather domains, by progressively learning a new domain at a time while retaining the previous learned knowledge. Our method conceives three novel concepts: adaptive knowledge acquisition, pseudo-label blending, and weather composition replay.

In contrast to single-target unsupervised domain adaptation, our sequential domain adaptation aims to both learn from the new target and retain previously acquired knowledge simultaneously.
Because of this, our model needs to identify potentially detrimental input regions that could introduce significant domain gaps and lead to forgetting of previously learned knowledge \cite{yang2022st++, kalb2023principles}.
To achieve this, we introduce adaptive knowledge acquisition by utilizing the previous model and class-wise feature representations, resulting in a dynamic weighting map.
This dynamic weighting map acts as a constraint, preventing the current model from learning potentially detrimental areas.

Various adverse weather conditions may exhibit similar degradation effects~\cite{kalb2023principles,li2023vblc}.
For instance, both fog and the rain veiling effect visually look alike.
Based on this similarity, models trained under different adverse weather conditions can collaborate to enhance learning from the shared degradation patterns~\cite{allen2020towards}.
Motivated by this idea, we propose a pseudo-label blending strategy. This involves employing the previous model as an auxiliary model to identify images from the new target that share similarities with those from the past targets. We then involve the auxiliary model (i.e., the previous model) to enhance our learning process in an ensemble manner. Note that, since we use a teacher-student framework, the term "previous model" refers to the previous teacher.

When operating in a sequential manner, we do not assume accessibility to the images from previously learned targets.
In such a scenario, to maintain the previously acquired knowledge, we introduce a replay technique that serves as a continuous reminder of the weather degradation patterns encountered in the past.
To implement this, we retain the weather information acquired from different targets in the previous steps, 
Once a new target is introduced, for each target image, we randomly augment each of the past weather information instances and integrate them into random segments of the current target images.
Through training on these composite images, the model sustains and revises its comprehension of various weather degradations over time, even when direct access to previously learned target domain images is unavailable.
In a summary, our contributions are as follows:
\begin{itemize}
	\item 
	We present an adaptive knowledge acquisition method that guides the model to refrain from learning new contents that could potentially result in forgetting.
	\item  
	We introduce the concept of incorporating the previous model into the current learning process to enhance our method's overall performance in an ensemble manner.
	\item
	To retain the past weather information, we propose a continuous replay of previously learned weather degradation, by randomly augmenting and integrating it into the present target images.
\end{itemize}
Our method consistently outperforms the state-of-the-art methods, and obtains the best performance with averaged mIoU (\%) of {\bf 65.7} and the lowest forgetting (\%) of {\bf 3.6} against 60.1 and 11.3 \cite{hoyer2023mic}, on the ACDC datsets for a four-target continual multi-target domain adaptation.

\section{Related Work}
\label{sec:relatedwork}

Unsupervised domain adaptation (UDA) has been explored extensively in recent years, with many applications ranging from different vision tasks~\cite{ganin2015unsupervised,chen2018domain,chen2021scale,vu2019advent,saito2019strong,zou2018unsupervised,li2019bidirectional}.
However, UDA settings are limited to one source and one target, meaning that the trained domain adaptive model can only work on a certain target domain and will fail if more target domains are involved.
Hence, some researchers start to explore methods to adapt a model into multiple target domains~\cite{peng2019domain,chen2019blending,yu2018multi,gholami2020unsupervised,nguyen2021unsupervised,yao2022federated,roy2021curriculum}.

\cite{isobe2021multi,saporta2021multi,lee2022adas} adapt the model into multi-target domains in a parallel way, where all the target domains are involved in every iteration.
\cite{isobe2021multi} maintains an expert model for each target domain, they are trained on many augmented images and teach a common student model via a knowledge distillation loss.
\cite{saporta2021multi} proposes two MTDA methods, Multi-Dis and MTKT.
For each target, Multi-Dis uses two types of domain classifiers,  a source vs. target classifier and a target vs. all the other targets classifier.
MTKT has a target domain-specific decoder, and a corresponding target-specific domain classifier for each target domain, the knowledge learned from different decoders is passed to a common decoder with knowledge distillation. 
\cite{lee2022adas} disentangles the input images into semantic contents and style contents, they obtain source images in target styles by swapping the contents.
The drawbacks of the parallel approaches are, (1) They need additional modules/images for every target domain in every iteration, and the real-time memory consumption could be a bottleneck when the number of target domains becomes larger. (2) Similar to the single-target UDA model, when a new target domain is introduced, there is no way to update a pretrained model, and hence a new model will need to be trained from scratch with all the targets.

\cite{saporta2022multi} proposes a continual way to adapt a model to multiple target domains, where the target domains are adapted through multiple steps, and the model can be updated at any time when a new target domain is introduced.
The paper also points out that due to this problem, the continual approach usually performs worse than the parallel approach.

\textit{Replay} has been proven effective in addressing this preserving previously learned knowledge for adverse weather conditions \cite{kalb2023principles}. 
Existing replay techniques including storing representative exemplar images \cite{rebuffi2017icarl,hayes2020remind,kang2022class} and adversarially generating images in the style of previous domains \cite{shin2017continual}.
However, the weather degradation is closely coupled to the image's physics properties (e.g., depth) \cite{sindagi2020prior,hu2021single,yang2022self,li2023vblc}, so the degradation in each image is unique and it is hard to identify a representative exemplar image.
Moreover, generating synthetic weather degradation is still an unsolved problem \cite{sakaridis2018semantic,anoosheh2019night}, and there does not exist a solution to continuously generate different adverse weather conditions.

\begin{figure*}
	\centering
	\includegraphics[width=\textwidth]{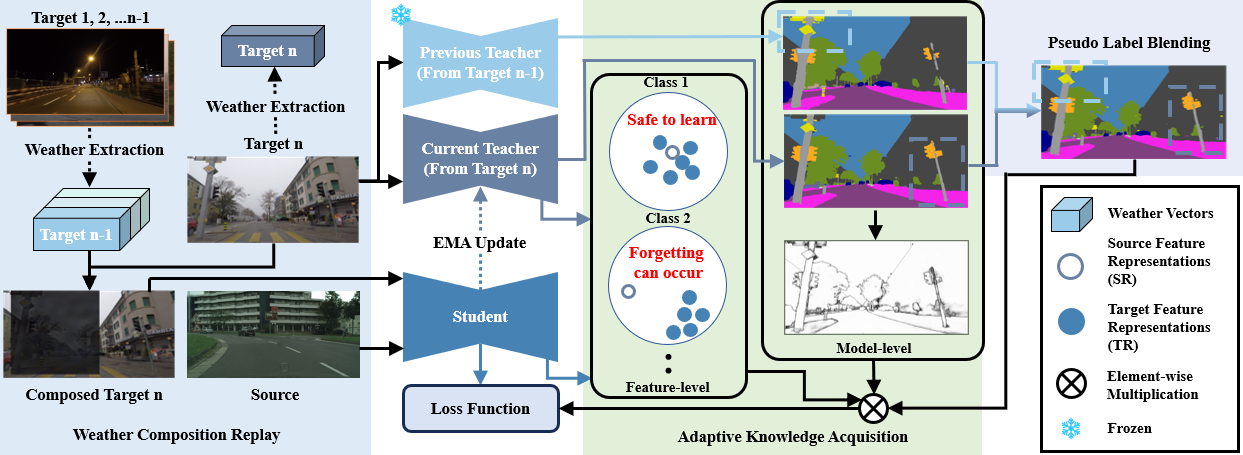}
	\caption{\label{fig:architecture}
		Our architecture for adapting a model to $n$ adverse weather conditions in $n$ steps in a sequential manner. The architecture consists of several key components: (1) Adaptive knowledge acquisition, where the model is guided to avoid learning the areas that could lead to a forgetting problem. (2) Pseudo-label blending, where the previous teacher is involved for enhancing the pseudo-label. (3) Weather composition replay, where the weather vectors from previous steps are composed into the current target image for revising on previously learned knowledge.
	}
\end{figure*}

\section{Proposed Method}
Our goal is to adapt our model sequentially to multiple adverse weather conditions, focusing on one weather domain at a time, all the while retaining the knowledge gained previously.
This sequential learning is necessary since we do not assume that once we have learned a domain, the data from that particular domain is accessible.
To achieve this goal, we begin by adapting our model to an initial weather domain.
Subsequently, we utilize the knowledge gained from the first domain (previous domain), which includes the teacher model and weather vectors, to aid the model's adaptation to the second weather domain (current domain).
This process of acquiring and retaining knowledge is iterated for each weather domain.

Fig.~\ref{fig:architecture} shows our pipeline. 
We progressively adapt our model to $n$ distinct adverse weather domains over $n$ steps.
We feed the source images to the student model in each step for supervised learning on semantic segmentation.
As for the target parts, from step $1$ to step $n-1$, our weather composition replay extracts and preserves weather vectors for every previously encountered target domain.
Then, in step $n$, we compose the extracted weather vectors into the current target image.
This composite image is subsequently fed into the student model for making predictions.

We combine pseudo-labels generated by both the current teacher model and the previous teacher model to synergistically improve the quality of pseudo-labels.
Our adaptive knowledge acquisition assesses pixel-wise domain shifts from both model-level and feature-level viewpoints. This dynamic reassessment enables us to adjust the learning process to the current target image, helping the model avoid incorporating detrimental information that may result in significant forgetting.

\subsection{Adaptive Knowledge Acquisition}
In this process of adaptive knowledge acquisition, we identify image regions with significant domain shifts that could potentially trigger model forgetting. Subsequently, we dynamically adjust the weighting of these areas, ensuring that while acquiring new knowledge, the model can still retain past knowledge.

\subsubsection{Model-Level Adjustment}

Following the standard unsupervised domain adaptation methods~\cite{kennerley20232pcnet,hoyer2023mic},  
in each step $n$ (weather domain $n$), 
we obtain a teacher model through the EMA (Exponential Moving Average) of a student model. 
We then adapt the student model to the new target domain by learning from the soft pseudo-labels generated by the teacher.

In this standard process, the student will quickly adapt to the new target domain. However, certain images in the new target domain can result in forgetting previous knowledge, especially for images that differ significantly from those in the previous target domain.
To mitigate this potential risk, we incorporate the teacher model from a previous target domain as an auxiliary model. This teacher model has not been exposed to the new target images during training. We refer to this teacher model as \emph{Previous-Teacher}.

Our Previous-Teacher makes a prediction for each image in the current step.
Consequently, the confidence of our Previous-Teacher in areas affected by the new adverse weather degradation can be used as an indication of the degree of the domain shift.
Since, its confidence is low for areas that have never been learned before.
This allows our model to selectively learn from the new target, avoiding harmful (low confidence) areas and thereby reducing the risk of forgetting previously acquired knowledge~\cite{kalb2023principles}.
To implement this idea, we define confidence as $q$:
\begin{equation}\label{eq:certainty}
	q(x)_{ij} =\max(g(x)_{ij}),
\end{equation}
where, $g$ is the model and $x$ is the input.

We compute a pixel-wise dynamic weighting mask for the model-level adjustment, denoted as $M^{\rm mod}$, as follows: 
\begin{equation}\label{eq:mm}
	M^{\rm mod}_{ij} =(1 - \alpha) q_{\rm cur}(x^T)_{ij}+\alpha q_{\rm pre}(x^T)_{ij},
\end{equation}
where $x^T$ represents the target image, $q_{\rm pre}$ and $q_{\rm cur}$ represent the confidence scores from the previous teacher and the current teacher, respectively.
$\alpha$ is a parameter to control the weights of each term, where it is large initially, and reduces along the number of iterations.
For regions where Previous-Teacher exhibits significant low confidence, $M^{\rm mod}$ dynamically reduces the learning weight in those areas. 
As a result, the model is encouraged to prioritize exploration in the regions that are less likely to lead to forgetting.
As the model progressively strengthens its resilience to new weather degradations, the confidence $q_{\rm cur}(x^T)_{ij}$ in these challenging regions increases. Simultaneously, with a decrease in the parameter $\alpha$, the model becomes more inclined to adaptively learn from these areas.

\subsubsection{Feature-Level Adjustment}
While model-level adjustment focuses on utilizing teacher models to guide the model's learning, feature-level adjustment aims to utilize class-specific feature representations to quantify the domain shift caused by weather degradation.

When a model attempts to adapt to a new adverse weather domain, the low-level features it extracts from the early convolutional layers are prone to degradation due to the new weather conditions. Consequently, predictions relying on these imprecise features could result in erroneous outcomes and induce substantial domain shifts within the feature space. In contrast, a model capable of accurately extracting information should avoid notable domain shifts in the feature space.
Therefore, we leverage this feature information to assess the extent of domain shifts.

To implement this, we first define the feature representations.
Based on the ground-truth of a source image, we partition the feature maps for different classes and calculate their average for each class, yielding class-wise source feature vectors.
We then compute the exponential moving average (EMA) of the source feature vectors, denoted as $\rm SR$.
Compared to the source feature vectors, $\rm SR$ can filter out images that are different from the typical source images, and thus can provide more representative features.

Similarly, we partition and calculate the average of the target feature maps according to the corresponding pseudo-labels, resulting in target feature vectors.
As our objective involves dynamically assigning weights to each target image, we directly employ the target feature vectors as the target feature representation, denoted as $\rm TR$, for every image.
Following this, we construct a weighting map based on the distances between $\rm SR$ and $\rm TR$:
\begin{equation}\label{eq:mf}
	M^{\rm feat} = \max(0,1-\frac{({\rm TR}_{c_1}-{\rm SR}_{c_1})^2}{\sum_{c=1}^{C}({\rm TR}_{c}-{\rm SR}_{c_1})^2}),
\end{equation}
where  $c_1$ represents the predicted class, and $C$ is the number of all the predicted classes.
In this equation, we calculate the relative distance between $\rm SR$  and $\rm TR$ for class $c_1$.

Intuitively, as shown in Fig.~\ref{fig:architecture}, within the feature space, for the target feature representations of Class 1, if they closely resemble the source feature representations of the same class, the domain shift is probably minor.
This implies that learning in these areas would result in lesser chances of significant forgetting.
Conversely, when we notice that the target feature representations are considerably distant from the source feature representations of the same class (as illustrated in Class 2 in the figure), it indicates that the model might have captured degraded information from the image.
Training within these regions can lead to substantial alterations in the feature space, potentially affecting the pretrained feature structures~\cite{hoyer2022daformer, su2023neighborhood}.
In such scenarios, our objective is to mitigate the possibility of forgetting by constraining the model from learning based on this information.%

\subsection{Pseudo-Label Blending} 

By integrating Previous-Teacher with the current teacher model, we can effectively ensemble them to enhance the robustness of the pseudo-labels~\cite{allen2020towards}.
We achieve this using both Previous-Teacher and its target feature representations.
First, we utilize both the Previous-Teacher and the current teacher model to generate predictions for the current target image.
Based on the confidences of these predictions, we generate a binary mask, denoted as $M^{\rm con}$, which indicates the regions where the Previous-Teacher exhibit higher confidence compared to the current teacher,

\begin{equation}
	M^{\rm con}_{ij}= 
	\begin{cases}
		1,				& \text{if } q_{\rm pre}(x^T)_{ij} > q_{\rm cur}(x^T)_{ij}\\
		0,              & \text{otherwise}
	\end{cases},
\end{equation}
As discussed in the feature-level adjustment, if Previous-Teacher demonstrates greater robustness to an area than the current teacher for a specific image segment, the corresponding extracted target feature representations should be closer to the source feature representations.
Hence, we also compute a weighting map using the equation mentioned in Eq.~(\ref{eq:mf}), but using the target feature representations generated from Previous-Teacher.
We denote this weighting map as $M^{\rm feat}_{\rm pre}$.
We can then obtain the refined pseudo-label, denoted as $p$.
\begin{equation}
	p(x^T)_{ij} = \operatorname*{argmax}_c (q_{\rm cur}(x^T)_{ijc} + M^{\rm con}M^{\rm feat}_{\rm pre}q_{\rm pre}(x^T)_{ijc}).
\end{equation}
We incorporate reliable predictions from the previously learned models into the current learning process. 
This blending of pseudo-labels enables the model to effectively learn the similar patterns in different targets, leveraging the knowledge and expertise accumulated by both Previous-Teacher and the current teacher model.

\subsection{Weather Composition Replay}
\begin{figure}
	\centering
	\begin{subfigure}[b]{.495\columnwidth}
		\centering
		\includegraphics[width=\linewidth,height=0.6\linewidth]{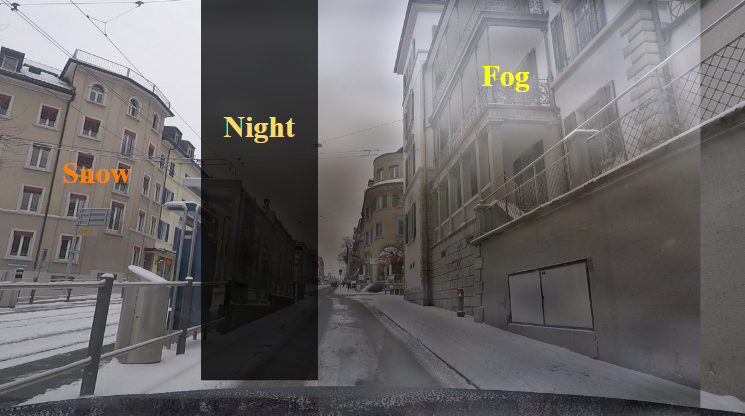} 
	\end{subfigure}
	\begin{subfigure}[b]{.495\columnwidth}
		\centering
		\includegraphics[width=\linewidth,height=0.6\linewidth]{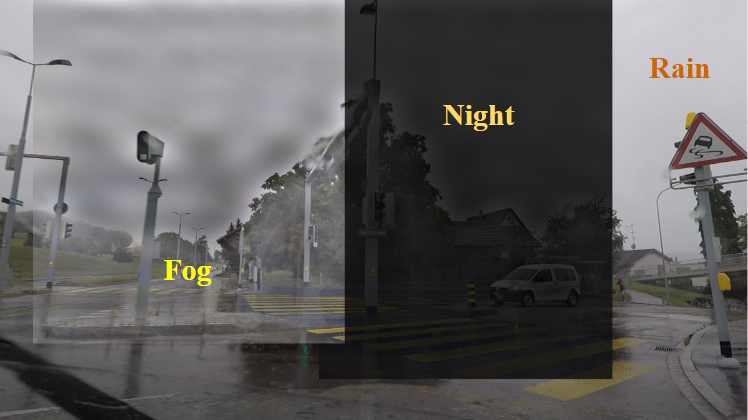} 
	\end{subfigure}
	\caption{\label{fig:compose} 
		Examples of composing night and fog weather vectors into snow and rain images, respectively.}
\end{figure}

We propose a replay technique utilizing the weather information obtained from different weather domains. 
As mentioned in \cite{kalb2023principles}, the domain shifts introduced by weather degradation can be captured in the frequency domain by the amplitude spectrum.
Hence, when adapting to a new target domain, we begin by translating each new target image into the frequency domain, and separate the amplitude from the phase.
Although each target image may contain some image-specific information, they are all subjected to the same adverse weather conditions. 
By averaging the extracted amplitude, we can then suppress image-specific information while preserving the weather vectors, in frequency domain.

As illustrated in Fig.~\ref{fig:architecture}, once the weather vectors are extracted, we keep them to future steps.
In step $n$, we apply augmentations on the stored weather vectors from step 1 to step $n-1$, and randomly inject them into the current target images as follow:
\begin{equation}
	x^{T2}_{r} = M_r x^{T2}+(1-M_r) {\rm iFT}(P^{T2},\sigma A^{T1}),
\end{equation}
where, $x^{T2}$ is the current target image, $x^{T2}_r$ is the composed images for the model to learn, $P^{T2}$ and $A^{T1}$ represent the phase component of the current image and the weather vectors of the previous domain, respectively. $\sigma$ represents a randomly determined volume of $A^{T1}$, where a larger volume indicates a stronger dominance of the previous weather information in the area. $\rm iFT$ represents an inverse Fourier Transformation to translate back the images from frequency domain. $M_r$ represents a random segment in that image, as augmenting a segment can be more efficient than augmenting the whole image \cite{yun2019cutmix,olsson2021classmix}.
This process is repeated for each weather vector we obtained, for a total of $n-1$ times.

This replay technique, utilizing the composed images, ensures that the model continuously updates its understanding of adverse weather conditions while retaining and refining its knowledge from past experiences.

\begin{table*}[t]
	\centering

		\begin{tabular}{lccccccc}
			\noalign{\hrule height 1.5pt}
			\multicolumn{8}{c}{\textbf{Cityscapes$\rightarrow$Night$\rightarrow$Rain$\rightarrow$Fog$\rightarrow$Snow}}                                                                                                               \\ \noalign{\hrule height 1.5pt}
			\multicolumn{1}{c|}{Method}           & \multicolumn{1}{c|}{\begin{tabular}[c]{@{}c@{}}Forgetting\\ Prevention\end{tabular}} & Night & Rain & Fog  & \multicolumn{1}{c|}{Snow} & \multicolumn{1}{c|}{\begin{tabular}[c]{@{}c@{}}mIoU\\ Avg. $\uparrow$ \end{tabular}} & \begin{tabular}[c]{@{}c@{}}A.F.\\  $\downarrow$ \end{tabular} \\ \noalign{\hrule height 1.5pt}
			
			\multicolumn{1}{l|}{MTKT \cite{saporta2021multi}}            & \multicolumn{1}{c|}{-}                                                               &   21.5 & 39.4  & 48.8 & \multicolumn{1}{c|}{38.7 } & \multicolumn{1}{c|}{ 37.1}      &        -      \\ \hline
			\multicolumn{1}{l|}{Multi-Dis \cite{saporta2021multi}}            & \multicolumn{1}{c|}{-}                                                               & 20.5     & 38.5 & 43.6 & \multicolumn{1}{c|}{36.8} & \multicolumn{1}{c|}{34.8}      &       -      \\ \hline
			\multicolumn{1}{l|}{AdvEnt \cite{vu2019advent}}           & \multicolumn{1}{c|}{\ding{55}}                                                               &   16.9 (-9.0)     & 36.9 (-3.2)  &  40.8 (-10.3) & \multicolumn{1}{c|}{35.1 } & \multicolumn{1}{c|}{32.4 }      &      22.4     \\ \hline
			\multicolumn{1}{l|}{MuHDi \cite{saporta2022multi}}            & \multicolumn{1}{c|}{\ding{51}}                                                               & 17.6 (-8.3)     & 37.2 (-3.4) & 44.4 (-6.7) & \multicolumn{1}{c|}{36.3} & \multicolumn{1}{c|}{33.9}      & 18.4            \\ \hline
			\multicolumn{1}{l|}{\textbf{Ours (DeeplabV2)}} & \multicolumn{1}{c|}{\ding{51}}                                                               & \textbf{24.0 (-1.9)}     & \textbf{42.0 (-1.3)} & \textbf{50.8 (-1.1)} & \multicolumn{1}{c|}{\textbf{44.0}} & \multicolumn{1}{c|}{\textbf{40.2}}      & \textbf{4.3}           \\ \noalign{\hrule height 1.5pt}
			\multicolumn{1}{l|}{MIC \cite{hoyer2023mic}}              & \multicolumn{1}{c|}{\ding{55}}                                                               & 34.7  (-7.2)   & 65.8 (-2.8) & 78.4 (-1.3) & \multicolumn{1}{c|}{65.2} & \multicolumn{1}{c|}{ 60.1}      &  11.3         \\ \hline
			\multicolumn{1}{l|}{\textbf{Ours (DAFormer)}} & \multicolumn{1}{c|}{\ding{51}}                                                               & \textbf{39.0 (-2.9) }    & \textbf{70.6 (-0.9)} & \textbf{80.4 (-0.2)} & \multicolumn{1}{c|}{\textbf{72.6}} & \multicolumn{1}{c|}{\textbf{65.7}}      &   \textbf{3.6}          \\ \noalign{\hrule height 1.5pt}
		\end{tabular}
	\caption{\label{tab:result3targets}Quantitative results of Ours compare to the existing unsupervised domain adaptation methods and continual multiple target domain adaptation methods, evaluated against four targets, ACDC nighttime, rain, fog and snow. Bold numbers are the best scores for different backbones. The mIoU (\%) of each target, the mIoU average of all the targets (the higher the better), and the accumulated forgetting, A.F. (the lower the better) are presented. The number in parentheses '()' indicates changes in performance, with a smaller number indicating a more pronounced forgetting effect. Our method outperforms the best existing method with forgetting prevention on DeeplabV2 backbone, by $6.3$ mIoU (\%) in average across all the targets, and $14.1$ in the accumulated forgetting. As for the DAFormer backbone, our method outperforms the best domain adaptation method by $5.6$ mIoU (\%) in average across all the targets, and $7.7$ in the accumulated forgetting.}
\end{table*}

\section{Experimental Results}

In this section, we present a comprehensive evaluation of our method under multiple adverse weather conditions in a sequential setting.
We begin by describing the datasets used in our experiments, followed by the architectures and parameters employed.
Next, we assess our model's performance both quantitatively and qualitatively, demonstrating its effectiveness in handling diverse weather conditions.
Lastly, we conduct ablation studies to evaluate the importance of each factor in our method.

\paragraph{Datasets}
We utilize Cityscapes \cite{cordts2016cityscapes} as our source domain, consisting of real-world street scene images captured under daytime, clear weather conditions.
For the target dataset, we employ ACDC \cite{sakaridis2021acdc}, which contains real-world street scene images captured under four adverse weather conditions: nighttime, rain, fog, and snow.
In our experiments, each of these four adverse weather conditions is treated as a separate target, and our models are adapted to these targets sequentially.

\paragraph{Baseline Models}
In our experiments, we conduct comparisons with the state-of-the-art unsupervised domain adaptation method, MIC \cite{hoyer2022daformer}.
To ensure a fair comparison, we use the same architecture, DAFormer, in all comparisons. Additionally, we employ identical optimization strategy, including the number of epochs, batch sizes, domain adaptation techniques and the pretrained backbone, as suggested in MIC.

Since our method is not limited by the model's architecture, we also compare it with two parallel and one continual general-purpose multiple target domain adaptation methods, MTKT, Multi-Dis, and MuHDi \cite{saporta2021multi,saporta2022multi}. 
Note, for the parallel setting methods \cite{saporta2021multi}, they learn all the targets simultaneously, their models are not affected by the forgetting problem.
For the continual setting method \cite{saporta2022multi}, it provides a forgetting prevention technique, but this method is not designed specifically for adverse weather conditions.
Once again, to maintain fairness in the comparison, we utilize the same architecture, DeeplabV2 \cite{chen2017deeplab}, and employ identical optimization strategy, domain adaptation techniques and pretrained backbone, following \cite{saporta2022multi}.

Regarding our method's parameters, we initialize the value of $\alpha$ to be $0.8$ and gradually decrease it to $0.2$ as the number of iterations progresses.
When injecting the previous weather information into the new target image, we randomly generate $\sigma$ between $0.2$ and $1.2$, and the sizes of the affected area are randomly selected in the target image, ranging from one-third to half of the image size.
All these parameters are decided empirically.

\paragraph{Evaluation Metrics}
Cityscapes and ACDC have the same segmentation class protocols, hence we apply the same protocol in our evaluation.
We use the percentage of Intersection over Union (IoU \%) as our evaluation metric for the effectiveness of the knowledge acquisition from a new target, the higher the better ($\uparrow$).
As for the knowledge retention, we use \textit{Accumulated Forgetting}, which is calculated as follow, 
\begin{equation}
	A.F. = \sum_{k=1}^{K-1} (mIoU_{k,k} - mIoU_{k,K}),
\end{equation}
where, $K$ is the number of targets.
We adapt the model to $K$ targets in $K$ steps.
$mIoU_{k,k}$ represents the initial performance of target $k$ at step $k$, and $mIoU_{k,K}$ represents the final performance of target $k$ at the last step, $K$.
A smaller Accumulated Forgetting indicates a lesser degree of forgetting in the model ($\downarrow$).

\subsection{Quantitative Results}

We conduct experiments on four ACDC adverse weather conditions: nighttime, rain, fog, and snow.
As shown in Tab.~\ref{tab:result3targets}, our models outperformed other methods on all targets.
Specifically, our models surpassed AdvEnt \cite{vu2019advent} and MIC \cite{hoyer2023mic} by $8.2$ mIoU (\%) and $5.6$ mIoU (\%) in mIoU Avg., and $18.1$ and $7.7$ in accumulated forgetting, respectively.
The highest mIoU and the smallest forgetting across all previous targets indicate the effectiveness of our models' knowledge acquisition and retention when adapting to different targets.

We compared our method to MuHDi \cite{saporta2022multi}, which offers a general-purpose forgetting prevention technique for multiple targets. Our method outperformed MuHDi by $6.3$ mIoU (\%) in mIoU Avg. and $14.1$ in accumulated forgetting.
Moreover, while the parallel multiple target domain adaptation methods do not suffer from the forgetting problem, our model still outperforms their performance in mIoU Avg. by $3.1$ mIoU (\%) and $5.4$ mIoU (\%), respectively.
This highlights the importance of our method's weather-specific knowledge acquisition and retention.

\begin{figure*}[ht!]
	\centering
	\begin{subfigure}[b]{.495\columnwidth}
		\centering
		\includegraphics[width=\linewidth,height=0.6\linewidth]{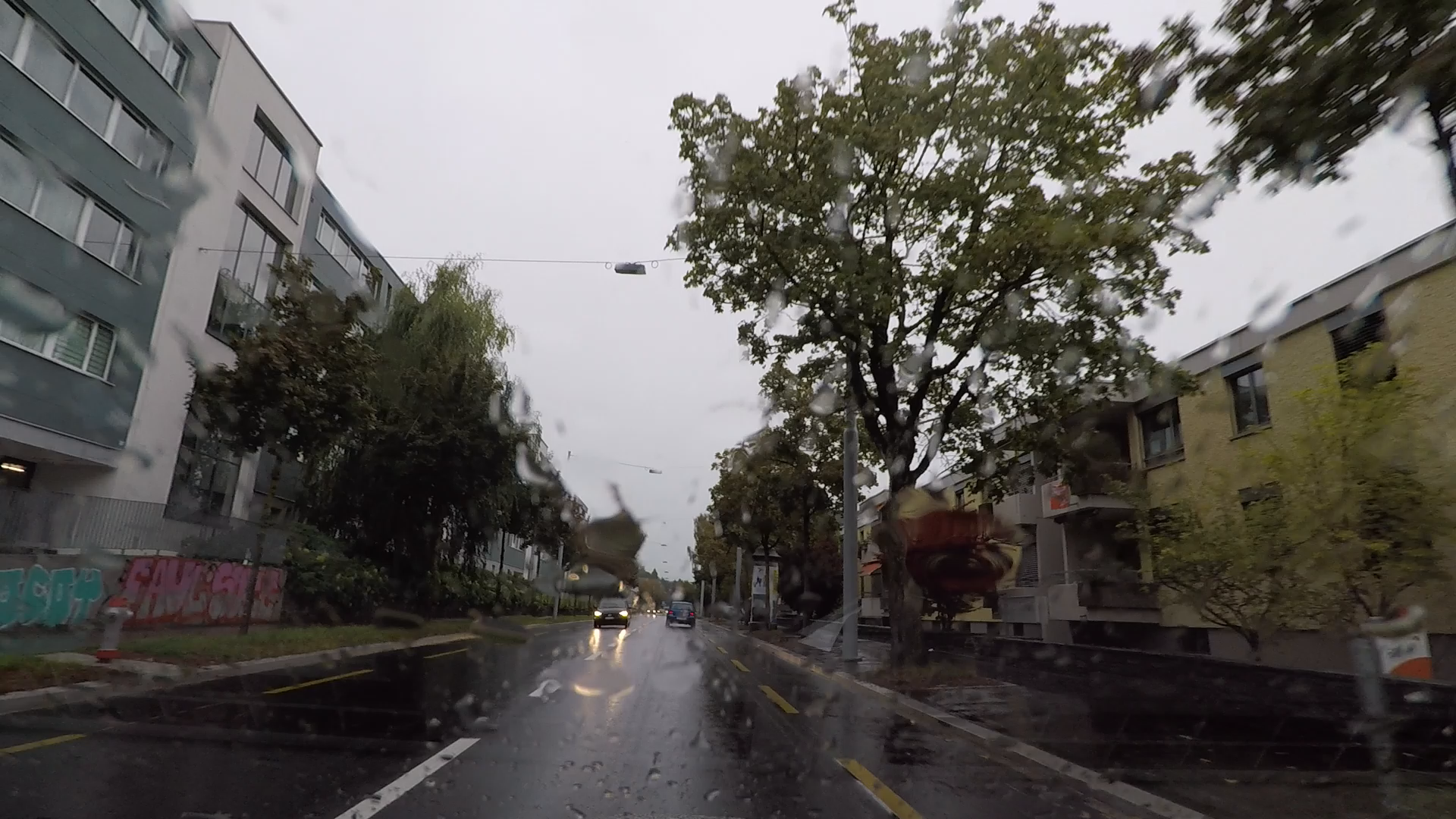} 
	\end{subfigure}
	\begin{subfigure}[b]{.495\columnwidth}
		\centering
		\includegraphics[width=\linewidth,height=0.6\linewidth]{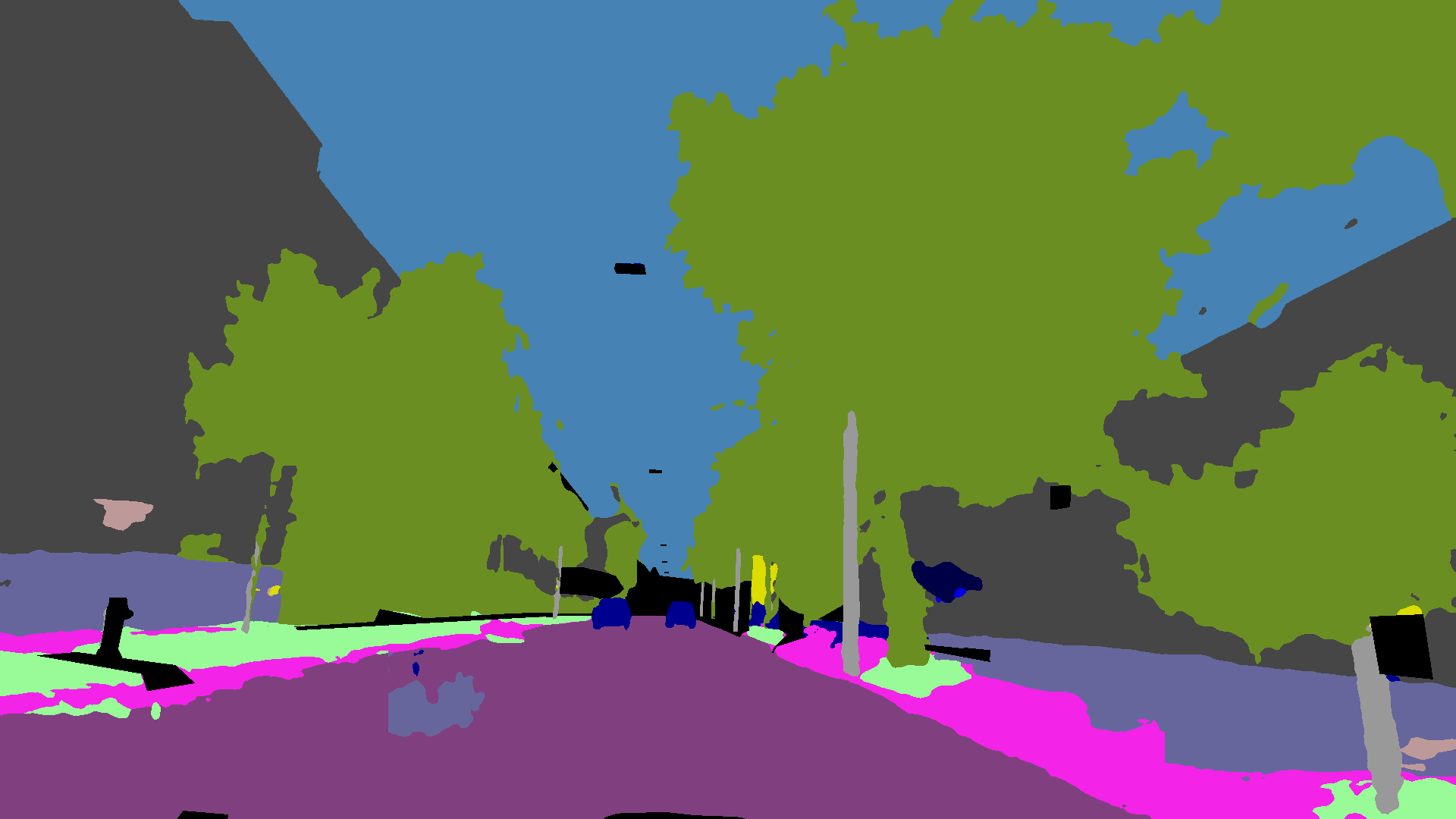} 
	\end{subfigure}
	\begin{subfigure}[b]{.495\columnwidth}
		\centering
		\includegraphics[width=\linewidth,height=0.6\linewidth]{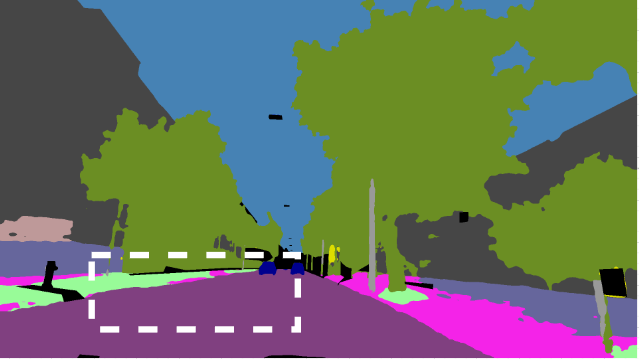} 
	\end{subfigure}
	\begin{subfigure}[b]{.495\columnwidth}
		\centering
		\includegraphics[width=\linewidth,height=0.6\linewidth]{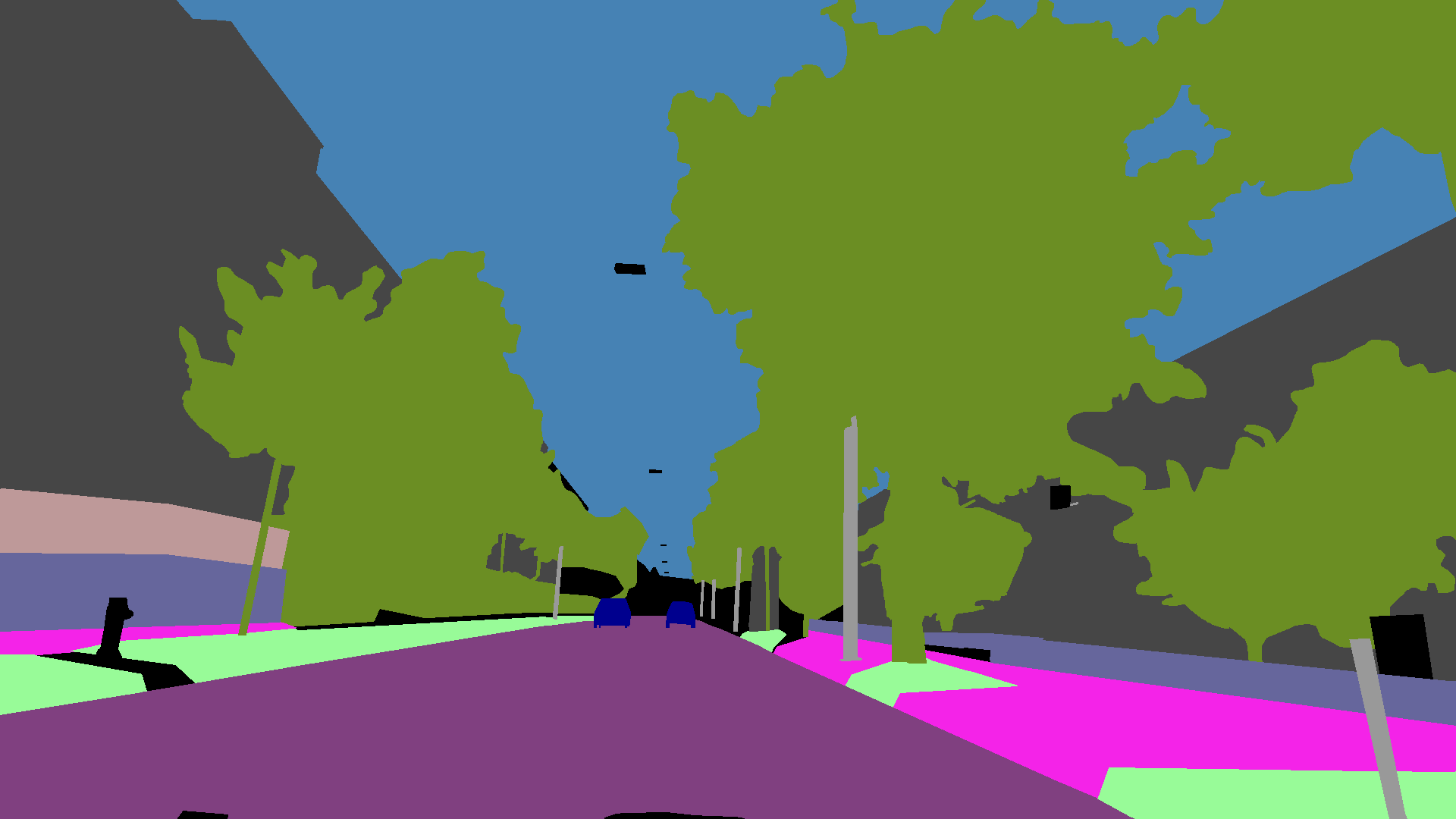} 
	\end{subfigure}
	\\
	\centering
	\begin{subfigure}[b]{.495\columnwidth}
		\centering
		\includegraphics[width=\linewidth,height=0.6\linewidth]{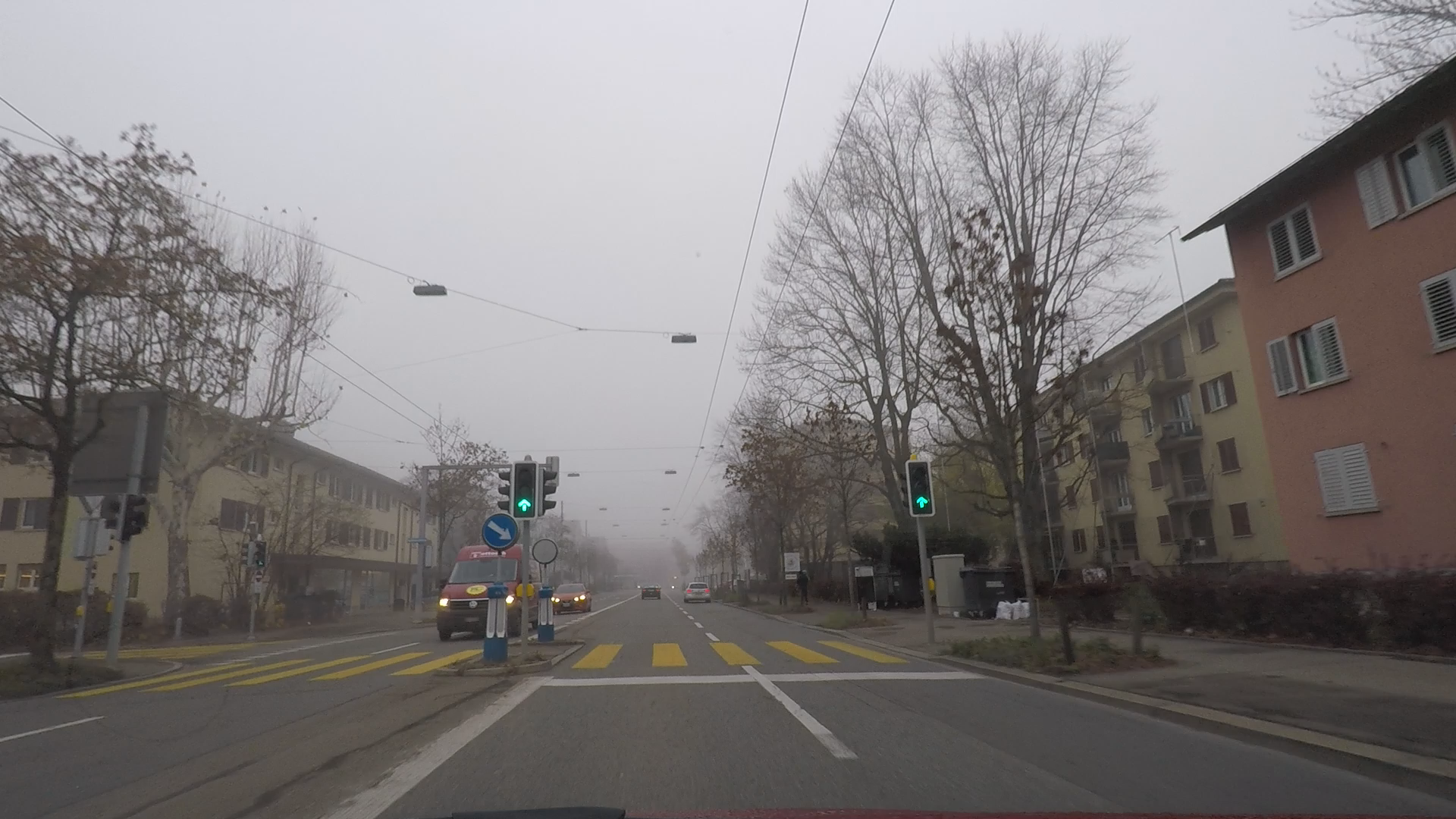} 
	\end{subfigure}
	\begin{subfigure}[b]{.495\columnwidth}
		\centering
		\includegraphics[width=\linewidth,height=0.6\linewidth]{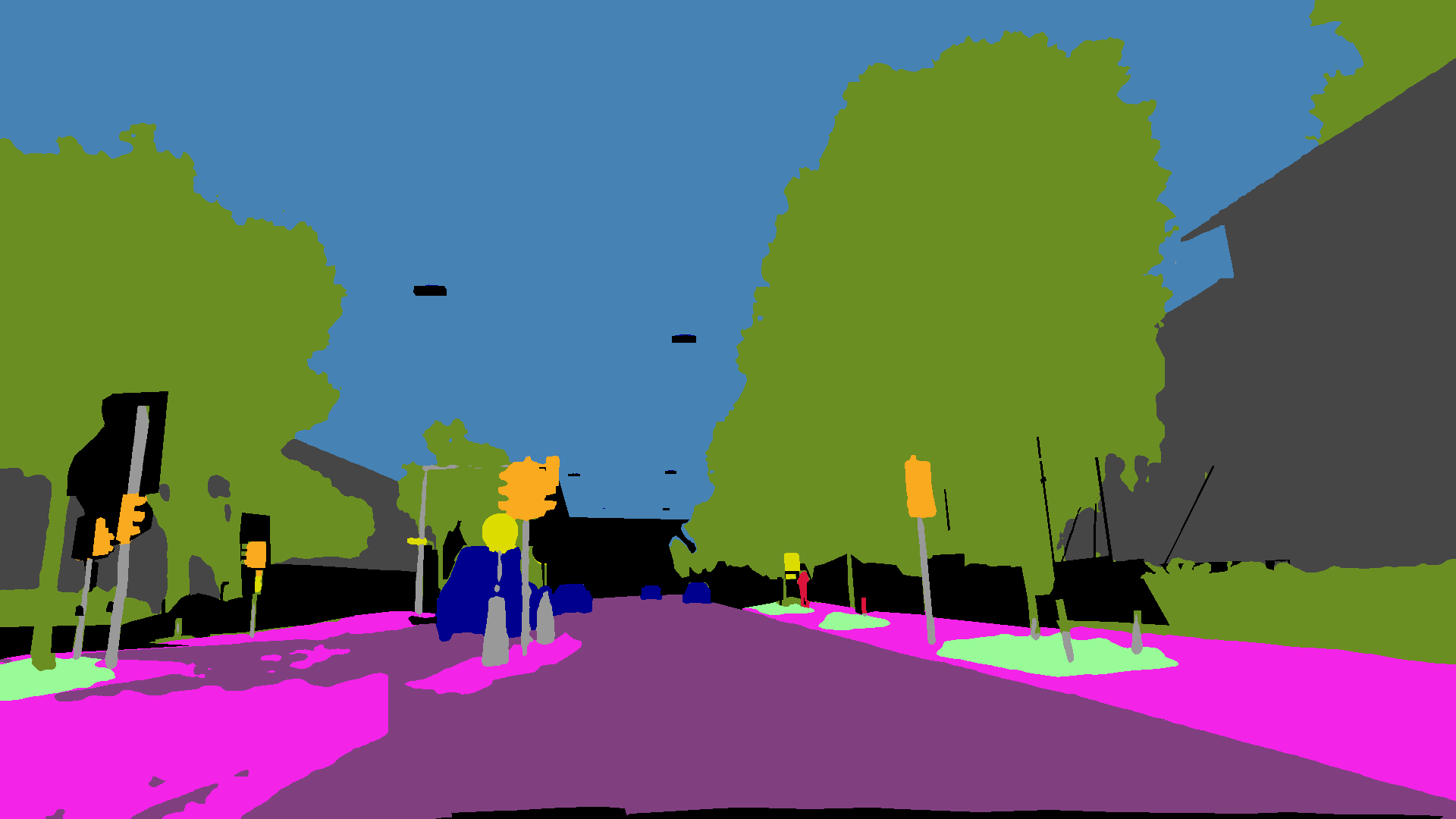} 
	\end{subfigure}
	\begin{subfigure}[b]{.495\columnwidth}
		\centering
		\includegraphics[width=\linewidth,height=0.6\linewidth]{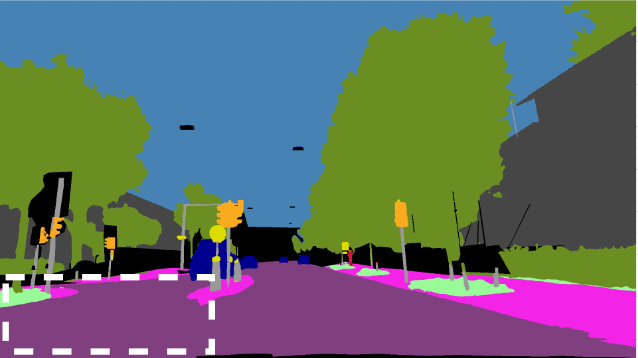} 
	\end{subfigure}
	\begin{subfigure}[b]{.495\columnwidth}
		\centering
		\includegraphics[width=\linewidth,height=0.6\linewidth]{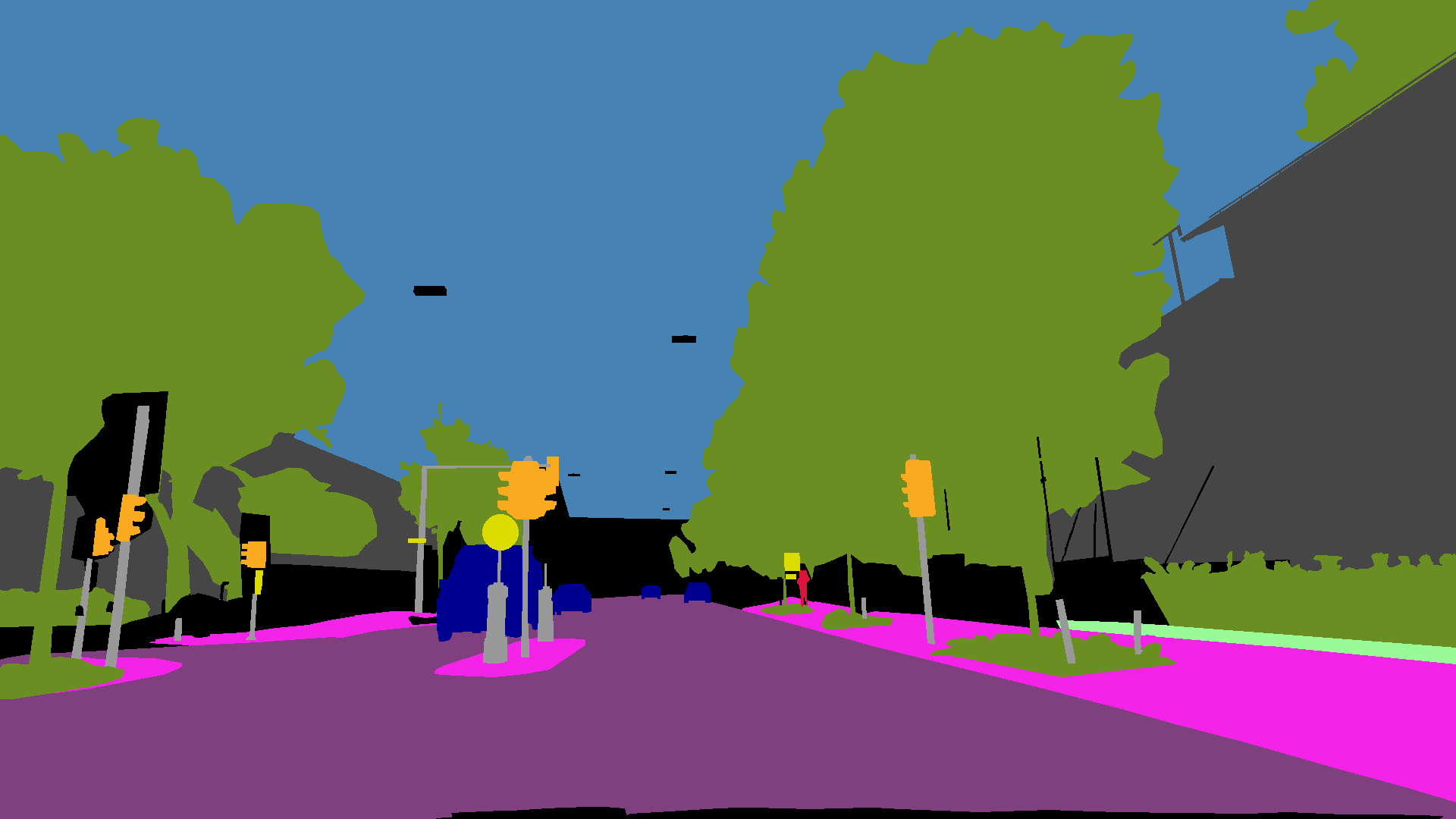} 
	\end{subfigure}
	\\
	\centering
	\begin{subfigure}[b]{.495\columnwidth}
		\centering
		\includegraphics[width=\linewidth,height=0.6\linewidth]{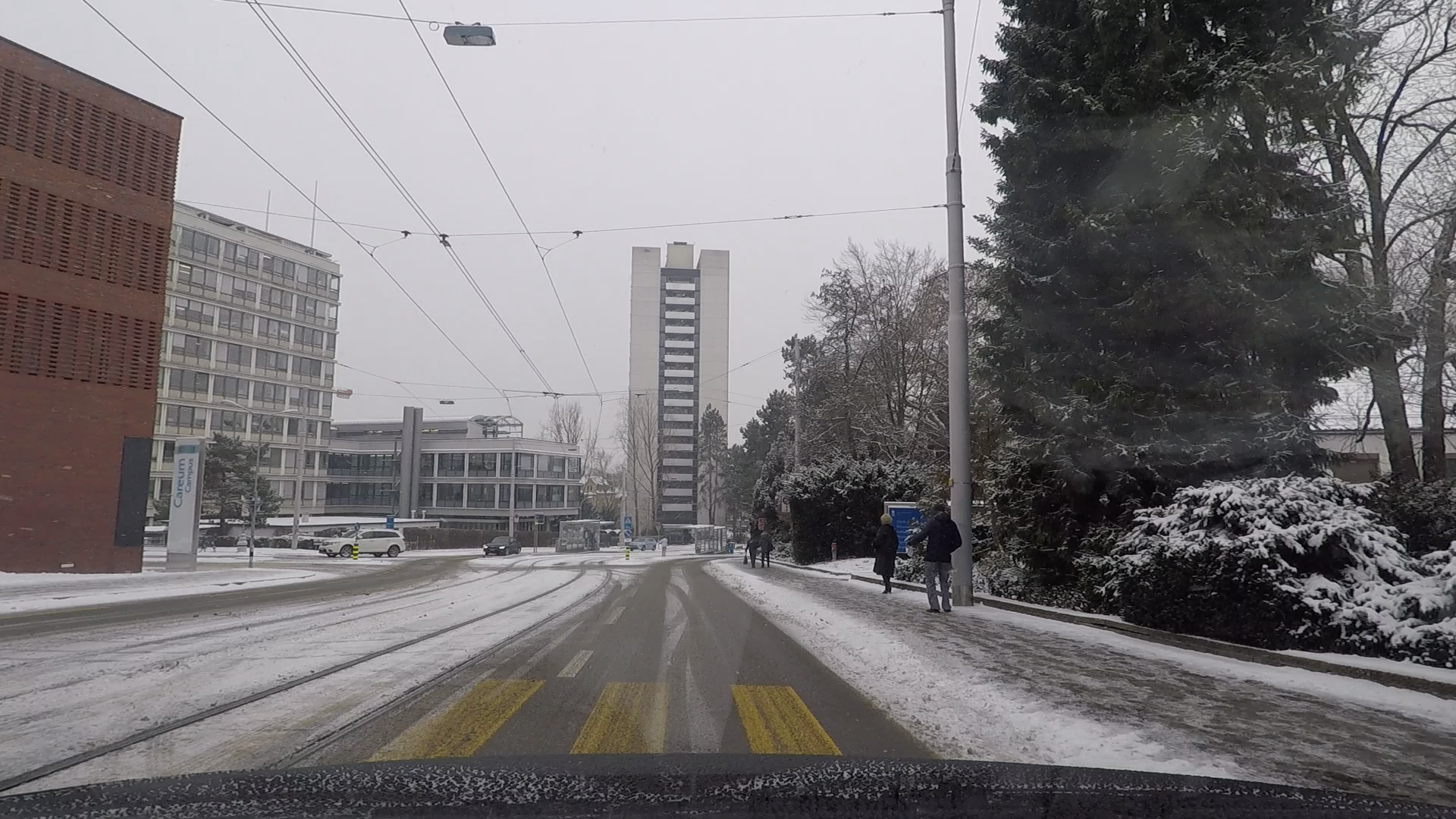} 
	\end{subfigure}
	\begin{subfigure}[b]{.495\columnwidth}
		\centering
		\includegraphics[width=\linewidth,height=0.6\linewidth]{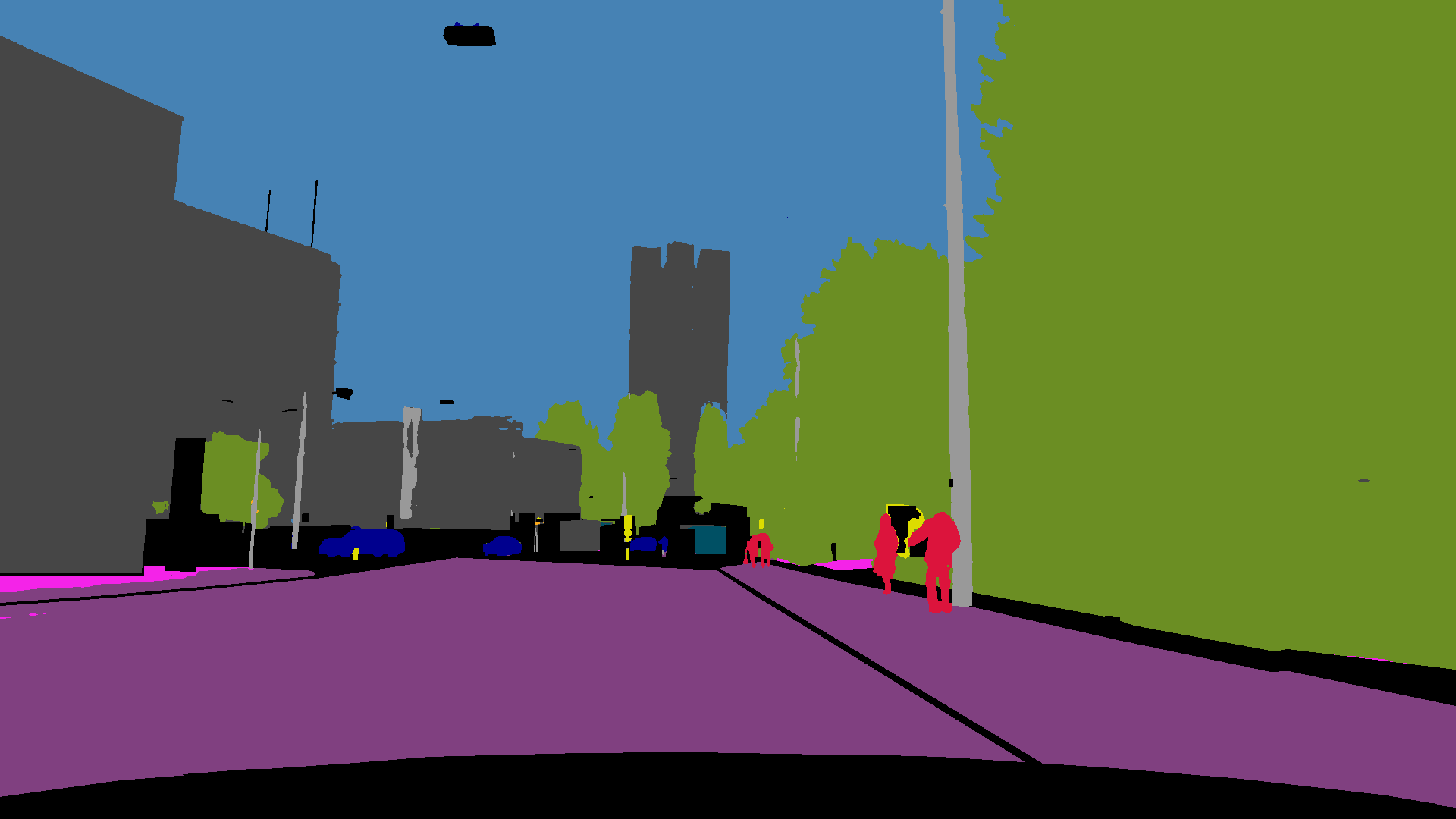} 
	\end{subfigure}
	\begin{subfigure}[b]{.495\columnwidth}
		\centering
		\includegraphics[width=\linewidth,height=0.6\linewidth]{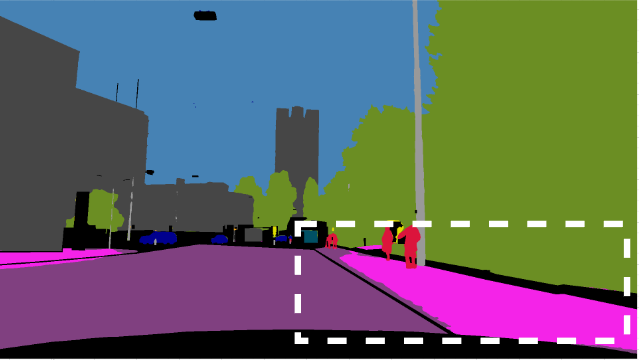} 
	\end{subfigure}
	\begin{subfigure}[b]{.495\columnwidth}
		\centering
		\includegraphics[width=\linewidth,height=0.6\linewidth]{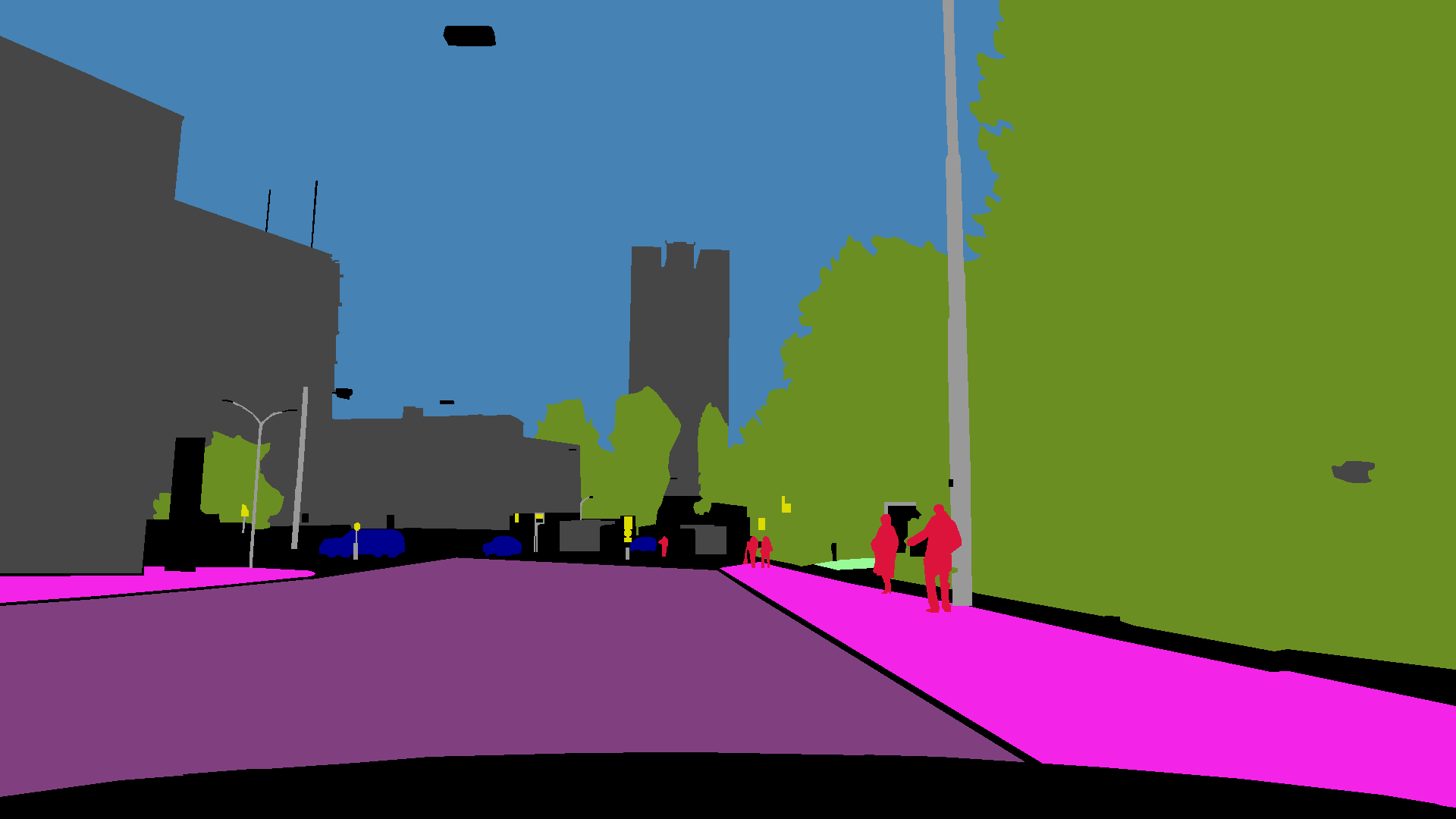} 
	\end{subfigure}
	\\
	\begin{subfigure}[b]{2\columnwidth}
		\centering
		\includegraphics[width=\linewidth]{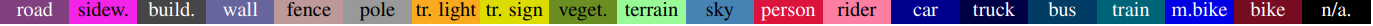} 
	\end{subfigure}
	\\
	\caption*{\ \ \ \ \ \ \ \ \ \  Image \ \ \ \ \ \ \ \  \ \ \ \ \ \ \ \ \ \ \ \ \ \ \ \ \ \ \ \ \ \ \ \ \ \ MIC \shortcite{hoyer2023mic} \ \ \ \ \ \ \ \ \ \ \ \ \ \ \ \ \ \ \ \ \ \ \ \ \ \ \ \ \ \ \ \ Ours \ \ \ \ \ \ \ \ \ \ \ \ \ \ \ \ \ \ \ \ \ \ \ \ \ \ \ \ \ \ \ \ \  Ground Truth \   \ \ \ \  }
	\caption{\label{fig:qualiresults}Comparisons on the semantic segmentation performance with MIC \cite{hoyer2023mic}, Ours (DAFormer), and ground truths on ACDC (Val.) under rainy, foggy, and snowy weather conditions following sequential multitarget domain adaptation. }
\end{figure*}

\paragraph{Without Source}
In certain circumstances, access to the source data may not be available in different steps.
In this section, we evaluate our method when we can only access the current target domain in each step. 
Since the source datasets are inaccessible, we do not apply $M^{\rm feat}$ and $M^{\rm feat}_{\rm pre}$ in this scenario.
We use MIC as the baseline for comparison.
Note, for both MIC and our method without source, we provide a model finetuned on the source, following \cite{kalb2023principles}.
Both methods are required to adapt this model to different adverse weather conditions.
For AdvEnt backbone, it requires the source dataset for domain adaptation, so models with this backbone are not involved.

The results are presented in Tab.~\ref{tab:forgetting}, with the absence of the source image dataset, MIC's accumulated forgetting increased significantly from $11.3$ (\%) to $23.9$ (\%), by $12.6$ (\%), where our model is less affected, with only an increase of accumulated forgetting by $6.7$ (\%).

\begin{table}[t]
	\centering\fontsize{9}{11}\selectfont
	\begin{tabular}{lcccccc}
		\noalign{\hrule height 1.5pt}
		\multicolumn{7}{c}{\textbf{Cityscapes$\rightarrow$Night$\rightarrow$Rain$\rightarrow$Fog$\rightarrow$Snow}}                                                                                                           \\ 
		\noalign{\hrule height 1.5pt}
		\multicolumn{1}{c|}{Method}             & Night & Rain & Fog  & \multicolumn{1}{c|}{Snow} & \multicolumn{1}{c|}{\begin{tabular}[c]{@{}c@{}}mIoU \\ Avg.  $\uparrow$\end{tabular}} & \begin{tabular}[c]{@{}c@{}}A.F. \\  $\downarrow$ \end{tabular} \\ \hline
		\multicolumn{1}{l|}{MIC (w/o)}                & 30.8   & 63.0 & 74.5 & \multicolumn{1}{c|}{69.2} & \multicolumn{1}{c|}{57.5}                                                 &     23.9                                             \\ \hline
		\multicolumn{1}{l|}{Ours (w/o)}  & 36.7  & 69.8 & 75.9 & \multicolumn{1}{c|}{69.6} & \multicolumn{1}{c|}{63.0}                                                 & 10.3                                                  \\ 
		\noalign{\hrule height 1.5pt}
	\end{tabular}
	\caption{\label{tab:forgetting} Quantitative comparisons between MIC \cite{hoyer2023mic} and our method, both without source (w/o).}
\end{table}

\begin{table}[t]
	\centering
	\begin{tabular}{lcc}
		
		\noalign{\hrule height 1.5pt}
		\multicolumn{3}{c}{\textbf{Cityscapes$\rightarrow$Night$\rightarrow$Rain$\rightarrow$Fog$\rightarrow$Snow}}                                                                                                                                                                \\ 
		\noalign{\hrule height 1.5pt}
		\multicolumn{1}{c|}{Method}       &  \multicolumn{1}{c|}{mIoU Avg.  $\uparrow$} & A.F.  $\downarrow$\\ \hline
		\multicolumn{1}{l|}{AdvEnt}         &  \multicolumn{1}{c|}{32.4}                                                 & 22.4                                                 \\ \hline
		\multicolumn{1}{l|}{Blending} & \multicolumn{1}{c|}{34.0}                                                 & 19.6                                                \\ 
		\noalign{\hrule height 1.5pt}
	\end{tabular}
	\caption{\label{tab:ablation}Ablation studies of knowledge acquisition with pseudo-label blending.}
\end{table}

\begin{table}[t]
	\centering
	\begin{tabular}{cccc}
		\noalign{\hrule height 1.5pt}
		\multicolumn{4}{c}{\textbf{Cityscapes$\rightarrow$Night$\rightarrow$Rain$\rightarrow$Fog$\rightarrow$Snow}}  
		\\ \noalign{\hrule height 1.5pt}
		\multicolumn{2}{c|}{Progressive}                                                                                                                          & \multicolumn{1}{c|}{\multirow{2}{*}{Replay}} & \multirow{2}{*}{\begin{tabular}[c]{@{}c@{}}A.F.\\$\downarrow$ \end{tabular}} \\ \cline{1-2}
		\multicolumn{1}{c|}{\begin{tabular}[c]{@{}c@{}}Model\\ Level\end{tabular}} & \multicolumn{1}{c|}{\begin{tabular}[c]{@{}l@{}}Feature\\ Level\end{tabular}} & \multicolumn{1}{c|}{}                        & \multicolumn{1}{c}{}                                                         \\ \noalign{\hrule height 1.5pt}
		\multicolumn{1}{c|}{}                                                     & \multicolumn{1}{c|}{}                                                        & \multicolumn{1}{c|}{}                        &                                                         22.4                                                                       \\ \hline
		\multicolumn{1}{c|}{\ding{51}}                                                     & \multicolumn{1}{c|}{}                                                        & \multicolumn{1}{c|}{}       &                  \multicolumn{1}{c}{12.9}                                                                  
		\\ \hline
		\multicolumn{1}{c|}{\ding{51}}                                                     & \multicolumn{1}{c|}{\ding{51}}                                                       & \multicolumn{1}{c|}{}                        & 9.6                                                                                                                                   \\ \hline
		\multicolumn{1}{c|}{\ding{51}}                                                     & \multicolumn{1}{c|}{\ding{51}}                                                       & \multicolumn{1}{c|}{\ding{51}}                       &                                                                4.5                                                                      \\ \noalign{\hrule height 1.5pt}
	\end{tabular}
	\caption{\label{tab:ablation2}Ablation studies of our knowledge retention techniques, including adaptive knowledge acquisition and weather composition replay.}
\end{table}

\subsection{Qualitative Results}
We first show our qualitative results of our knowledge acquisition ability in Fig.~\ref{fig:qualiresults}, for a four-target sequential domain adaptation (on night, rain, fog, and snow).
We evaluate our best model Ours (DAFormer) against MIC, and the ground truths semantic segmentation maps.
We can see that our method can predict more precise semantic segmentation maps, and also has fewer false positives compared to MIC in all the target domains.
More qualitative results are provided in the supplementary materials.

\subsection{Ablation Studies}
\paragraph{Knowledge Acquisition and Retention}
We begin by evaluating the effectiveness of our adaptive knowledge acquisition and weather composition replay techniques, as they are specifically designed for both knowledge acquisition and knowledge retention.
In Tab.~\ref{tab:ablation}, we report the performance changes of four models: AdvEnt, AdvEnt with model-level assistance, AdvEnt with both model-level and feature-level assistance, and AdvEnt with adaptive knowledge acquisition and weather composition replay techniques.
All these models are trained to adapt to four adverse weather conditions using source data, and it is evident that each component contributes to preventing forgetting.

\paragraph{Pseudo-Label Blending}
Pseudo-label blending is designed for exploring the similar patterns among different targets to improve the knowledge acquisition on the current target.
We evaluate the knowledge acquisition with two approaches: AdvEnt and AdvEnt with pseudo-label blending.
The results are presented in Tab.~\ref{tab:ablation}.
We can observe that pseudo-label blending enhances the average mIoU across the four targets, without causing additional forgetting issues.

Combining all our proposed techniques, we achieve an enhanced performance on the new target, while retain the most knowledge from the previous targets.

\section{Conclusion}
We have proposed a novel method that adapts a model to multiple unlabeled adverse weather conditions sequentially.
We use both model-level and feature-level knowledge to assist the model avoid learning from the harmful contents from the new target image that can lead to a forgetting of the previously learned knowledge.
To support our current learning process, we have also proposed a method to involve the previously obtained model to jointly improving the pseudo-labels for the current target.
We propose a weather composition replay technique, which compose the previously learned weather information to the current target image, enabling the model learn from the current target image while revising the previously learned weather information.
We train two models using our method on both DeeplabV2 and DAFormer, respectively, to demonstrate that our method can be generalized to different architectures. 
We evaluate these models on several benchmark adverse weather conditions with different settings and found that our models outperform many methods in different settings.

\bibliography{aaai24}

\end{document}